# Improving Part-of-Speech Tagging for NLP Pipelines

Vishaal Jatav, Ravi Teja, Srini Bharadwaj
Cognitive Intelligence Group
RAGE Frameworks, Dedham, MA, USA

And

Venkat Srinivasan
Senior Advisor, Genpact

**Abstract**

This paper outlines the results of sentence level linguistics based rules for improving part-of-speech tagging. It is well known that the performance of complex NLP systems is negatively affected if one of the preliminary stages is less than perfect. Errors in the initial stages in the pipeline have a snowballing effect on the pipeline's end performance. We have created a set of linguistics based rules at the sentence level which adjust part-of-speech tags from state-of-the-art taggers. Comparison with state-of-the-art taggers on widely used benchmarks demonstrate significant improvements in tagging accuracy and consequently in the quality and accuracy of NLP systems.

## I. INTRODUCTION

Part of speech tagging is the basic step of identifying a token's functional role within a sentence and is the fundamental step in any NLP pipeline. Several methods and approaches have been employed for POS Tagging, with various levels of performance. These approaches can be broadly divided into Statistical approaches and Rule-based approaches.

Almost all tagger performance results reported in the literature are at the word token level. Many state-of-the-art methods report a high level of accuracy in POS tagging at the token level. For example, the Stanford POS Tagger [Klein, 2003], TnT POS Tagger [Brants, 2000] and SVMTool POS Tagger [Giménez, 2004] have reported their average *token-level* accuracies at over 96%. Token-level metrics do not give a practical measure of POS tagging quality, especially with respect to how well the output can enable a correct end-product for the pipeline. Besides, even in the remaining 4% some are more critical than others. There is clear recognition that for accurately measuring the quality of POS-tagging from the perspective of the end product of the pipeline, we need to measure the correctness of POS-tagging for the sentence as a whole [Manning, 2011].

In this paper, we provide results at sentence-level from an approach to refine POS tags using linguistics based rules. We find that our approach can improve the quality of POS-tagging significantly over state-of-the-art taggers.

### A. Not all POS errors are equal

In analyzing errors in POS tagging, we segment them into *critical errors* and *non-critical errors*. *Critical errors* are those that can change the semantic interpretation of an entire sentence, typically due to a assigning an entirely incorrect POS category to a word, for example a Plural Noun (NNS) incorrectly tagged as a Present Tense Verb (VBZ). This alteration in the semantics has a deleterious effect on all the subsequent steps in the NLP pipeline – e.g. Syntactic Parsing, Dependency Parsing, etc.

*Non-critical errors*, on the other hand, are those where the effect of the tagging error is local and have less significant consequences. For example, a Proper Noun (NNP) incorrectly tagged as a Common Noun (NN). In such cases, the immediate *syntactic* steps might be affected, but the high-level *semantic* analysis of the overall sentence remains the same.

**Figure 1** gives some examples of POS errors on the Penn Tree Bank-3 data set [PTB] and the RAGE Reuters Dataset, which is a set of 110 Reuters articles.

---

**Sentence 1:** Wall Street posts sharp gains, fueled by strong consumer data
**POS Tagger Output:** Wall/NNP Street/NNP posts/NNS sharp/JJ gains/NNS ./, fueled/VBN by/IN strong/JJ consumer/NN data/NNS
**Comments:** *posts* got incorrectly identified as a Noun.

**Sentence 2:** An accompanying record of paralanguage factors for the second example might also note a throaty rasp.
**POS Tagger Output:** An/DT accompanying/VBG record/NN of/IN paralanguage/NN factors/NNS for/IN the/DT second/JJ example/NN might/MD also/RB note/VB a/DT throaty/JJ rasp/NN ./.
**Comments:** *accompanying* for incorrectly identified as a Verb.

**Sentence 3:** New home sales jumps in the third quarter and exceeds existing home sales.
**POS Tagger Output:** New/JJ home/NN sales/NNS jumps/VBZ in/IN the/DT third/JJ quarter/NN and/CC exceeds/VBZ existing/VBG home/NN sales/NNS ./.
**Comments:** *New Home Sales* and *Existing Home Sales* are valid concepts in the Industry.

**Sentence 4:** One thing's for sure: There have been a ton of them, and greater beings than the editors of the Wall Street.
**POS Tagger Output:** One/CD thing/NN 's/POS for/IN sure/JJ :/: There/EX have/VBP been/VBN a/DT ton/NN of/IN greater/JJR beings/NNS than/IN the/DT editors/NNS of/IN the/DT Wall/NNS Street/NNS ./.
**Comments:** *Contraction* is incorrectly identified as a *Possessive ending*

---

**Fig 1:** Selected Part-of-speech tagging errors in Stanford POS Tagger from Penn Tree Bank-3 and RAGE Reuters 110 Dataset

- *Critical.* In *Sentence-1*, the tagger incorrectly identified the main verb of the sentence, "posts' as a noun (NNS); the main action of <someone> <posting> <something> is lost in the processing, thereby eliminating this crucial aspect of the sentence in all subsequent processing of this sentence.

- *Critical.* In *Sentence-2*, the tagger misidentifies an adjective modifier (*accompanying*) as a verb gerund (VBG) despite the grammatical constraint that a determiner (DT) can never precede a verb (VBG, in this case), thereby altering the parse tree of the sentence to an unrecoverable loss of semantics for the phrase *Paralanguage Factors Records*.

- *Non-Critical. Sentence-3* shows two POS Tag errors. *New Home Sales* and *Existing Home Sales* are ground concepts in the *Real Estate* industry. In subsequent steps of an NLU pipeline, that attempts to understand the meaning of this sentence, the adjectival interpretation of *New*, instead of a compound noun interpretation of *New Home Sales* will take some semantics away from the final interpretation. Same repercussions will be seen with the concept *Existing Home Sales*.

- *Non-Critical. Sentence-4* interprets the *'s* as a *possessive* construct than a valid and widely used *contraction* – taking away the main verb of the sentence fragment *One thing is for sure.*

**B. The Problem with conventional POS taggers**

Most POS taggers are statistical sequence labeling models, implemented using statistical methods like HMM [Stratos, 2016] [Lee, 2000] [Goldwater, 2007], SVM [Giménez, 2004] [Nakagawa, 2001], Graph-based [Biemann, 2006] [Garrette, 2013], Perceptron-based [Ma, 2013] [Ma, 2014], etc. aiming towards generalization of the training data. They suffer from the common drawbacks of statistical machine learning. Some of these are elaborated below.

**Corpus Size.** The size of the training corpus for a NLP problem poses a trade-off. Large training corpus limits the performance of the learning algorithm – machines could take several hours/days to converge on the learning parameters. On the other hand, given the high computational vastness of the natural language, anything smaller than a life-size corpus may lead to incomplete models. This would ideally need a world-wide-web size corpus to represent all the grammatical rules – generalizations and exceptions. This leads to a trade-off – size of the corpus vs efficiency of learning (in the context of data and computational power) vs effective demonstration of granularity of context of the computationally complex natural language.

**Unknown Words.** A practical corpus, say, the Penn Tree Bank Corpus [Mitchel, 1994] or the Brown Corpus [Francis, 1964] that are used to train these statistical POS taggers, introduces the problem of unknown words. Most taggers apply heuristics to guess the correct POS tags of unknown words with varying levels of success [Huihsin, 2005] [Haulrich, 2009] [Nakagawa, 2001]. The problem of unknown words has another limitation: that of domain specificity [Béchet, 2000]. Models trained on newswire or other generic datasets prove inadequate for, say, Legal and Pharmacovigilance domains. More unknown words imply more guessing of the tags, thereby resulting in more errors. Furthermore, there is an added complexity that most real-world problems cannot be confined to clearly distinct domains. For example, the natural language text for *Insurance Endorsements* may be significantly different than, say, *risk assessment* of an *insured entity*.

**Lack of Context.** Machine learning algorithms based on word patterns work by discovering generalizations of word occurrence patterns across the corpus. By converting word tokens into vectors, such approaches lose essential context [Collobert, 2011] [Gens, 2017] [Luong, 2013] [Koo, 2008]. For a highly polysemous language like English, context and rhetorical structure is critical to understand the sense in which concepts are used and the meaning the author intends to convey.

**Corpus Quality.** The statistical model is only as good as the quality of the learning data. Research has shown a lot of inconsistencies and errors in the manual annotation of some of the benchmark datasets and corpus [Mitchell, 1994]. These could be because of genuine human errors or inconsistent tagging resulting from inter-annotator disagreements. The error rates in annotation were found to be as large as 7% at a token-level within the Penn Tree Bank [Mitchell, 1994]. As we show in the next section, this could propagate in **as many as ~25% of sentences**. The risk of errors (as seen from a practical sense of solving a business problem) is further exacerbated when linguists (who are weak in domain knowledge) try to build the learning corpus for a nuanced problem and SMEs (whose time is more expensive and have very less training in creating corpuses) often contribute in building the corpus.

**Traceability and Tractability.** Finally, the lack of traceability in statistical machine learning algorithms is a bottleneck for accurately learning POS Tagging. Given that it is practically impossible to capture all the context, senses, and scenarios in a dataset without annotation errors within them, it would be difficult to understand what was learnt by a statistical learning algorithm and how to fix the errors committed by the machine.

These challenges usually, significantly affect the quality of learning, and ultimately the performance of the end system. Despite this, they remain popular and generally in the domain of data scientists and not SMEs. Most data scientists may not understand these limitations or perhaps ignore them choosing instead to focus on optimizing the model parameters and feature engineering in the resulting models. An absence of a framework that effectively utilizes an SMEs time and efforts for tractable and traceable learning of machines is a major reason the industry and community are still drawn to statistical machine learning.

## II. EVALUATION METHODOLOGY

A well accepted evaluation methodology in any statistical evaluation problem involves an annotated [labeled] dataset that is split into a train set, a development set, and a test set, where the development set is used to tune the parameters of the model. To avoid *jackknifing* or over-estimation errors because of these splits, [Efron, 1982] suggested a k-fold cross-validation. The k-fold cross-validation randomly and repeatedly splits the data into *k* training and test subsets. Each time, one of the *k* subsets is used as the test set and the other *k-1* subsets are put together to form a training set. Then the average error across all *k* trials is computed. This reduces any over-estimation error as it matters less how the data gets divided. Every data point gets to be in a test set exactly once, and gets to be in a training set *k-1* times. The variance of the resulting estimate is reduced as *k* is increased.

In addition, a *formative or intrinsic evaluation* will focus on evaluation at the POS-tagging level, whereas *summative or extrinsic evaluation* will evaluate the performance at a use-case level (sentiment analysis, summarization, etc. or even syntactic parsing or semantic roles analysis). Our focus in this paper is on the formative evaluation albeit the conclusions carry intact into the summative.

In any evaluation methodology, the most common metrics measured have been the precision %, recall % and the F-score. Precision is the correctness of the output of the system, whereas recall is a measure of how generalizable the results are. Precision is computed on the train dataset and Recall is computed on the test dataset. F-score is an algebraic mix of precision and recall and reflects the robustness of the system by highlighting the high-precision low-recall or a low-precision high-recall trade-off.

The precision, recall and F-scores can be computed at either a token-level, sentence-level or at a document-level. The effectiveness of the evaluation is much more at aggregated-levels, than at granular level – which is where the true, contextual interpretation and understanding of the text by the system can be measured. Higher the numbers at the aggregated levels, higher will be the effectiveness of the system being evaluated.

**Table 1:** POS Tagging Metrics for some common systems

| System | Languages/Tags | Accuracy |
|---|---|---|
| [Béchet, 2000] | NP Evaluation | 72.6 |
| [Brants, 2000] | All Tags | 96.7 |
| [Choi, 2012] | All Tags | 93.05 |
| [Church, 1988] | All Tags | 99.5 |
| [Dandapat, 2007] | All Tags | 88.41 |
| [Das, 2011] | Multi-lingual All Tags | 83.4 |
| [Dredze, 2008] | Islandic All Tags | 91.54 |
| [Duong, 2013] | Multi-lingual All Tags | 83.4 |
| [Giménez, 2004] | Multi-lingual All Tags | 96.46 |
| [Hajič, 2001] | Czech All Tags | 95.16 |
| [Haulrich, 2009] | All Tags | 96.37 |
| [Hepple, 2000] | All Tags | 97.35 |
| [Huihsin, 2005] | Multi-lingual All Tags | 93.7 |
| [Kim, 2003] | All Tags | 96.9 |
| [Lee, 2000] | All Tags | 97.93 |
| [Lee, 2004] | Korean All Tags | 92.01 |
| [Li, 2015] | Multi-lingual All Tags | 95 |
| [Nakagawa, 2006] | Multi-lingual All Tags | 76.34 |
| [Saharia, 2009] | Assamese All Tags | 85.64 |
| [Schnabel, 2014] | All Tags | 93.14 |
| [Silfverberg, 2014] | Multi-lingual All Tags | 97.24 |
| [Sogaard, 2013] | Multi-lingual All Tags | 95.64 |
| [Sujith, 2009] | 17/45 Tags | 96.8 |
| [Sujith, 2014] | All Tags | 88.11 |
| [Sun, 2012] | All Tags | 95.34 |
| [Toutanova, 2003] | All Tags | 97.1 |
| [van Halteren, 2001] | All Tags | 97.23 |
| [Zhang, 2014] | Multi-lingual 17/45 Tags | 96.8 |

**Table 1** enumerates the high accuracies as reported by some of the most common POS systems within the research community at token-levels.

### III. RAGE-AI PLATFORM'S HYBRID PART-OF-SPEECH TAGGING COMPONENT

RAGE (Rapid Application Generation Engine) AI Platform is a no-code, data-driven, knowledge-based Business Process Automation platform, consisting of 20 engines to model business processes with complete traceability, tractability and flexibility. One of the 20-engines is the Computational Linguistics engine that is implemented as a deep-linguistics driven Natural Language Understanding engine. The linguistic rule-driven POS Tagger is part of the NLU pipeline.

The architecture of the RAGE-AI Hybrid POS Tagger has a generic, statistical part-of-speech tagger at its core. However, an extensive set of linguistics based rules are applied as a pre-processing step before using the statistical tagger. Then the part-of-speech tags from the tagger are iteratively applied with over 100 linguistic POS-correction rules, until there are no grammatical inconsistencies. Grammatical inconsistencies are detected using an additional set of 40+ grammar rules.

### IV. COMPARISON AND RESULTS [PART I]

The evaluation was conducted on 3 datasets – *Penn Tree Bank 3*, *RAGE Reuters 110* and *RAGE PubMed 110*. The baseline is the Stanford PCFG parser [Klein, 2003] – version 3.7.0 retrieved from the Stanford NLP website. This baseline is compared against the RAGE AI Platform's [Srinivasan, 2017] Hybrid Part-of-Speech Tagger Component (version 14.3). Metrics are reported at a sentence-level (instead of at the token level).

We provide two sets of results – unrelaxed and relaxed.

The Penn Tree Bank is about a 4.5 million words dataset [Mitchell, 1994]. Average number of words in a sentence is 15 which yields about 300,000 sentences. Considering a 95% precision (of any available POS tagger), there would be about 225,000 incorrect tokens. Assuming almost equal number of critical vs non-critical errors, as many as 112,500 tokens could be bad POS errors. In the very worst case, all these errors could occur in different sentences – thereby making ~38% of the sentences lose their original semantics.

RAGE Reuters-110 is a fully part-of-speech annotated corpus of over 110 news articles from Reuters from more than 20 different sections [domain/industries/sectors] of the News website spanning from *Entertainment* and *Sports* to *Deals* and *Economy*. The dataset is composed of over 21,000 words and the POS Tags are annotated using the Penn Tree Bank part-of-speech tag set.

RAGE PubMed-110 is a fully part-of-speech annotated corpus of over 110 abstracts from PubMed, a library of citations of over 27 million biomedical literature, life sciences journals and books. The methodology used to curate this dataset is collection of abstracts from over 20 therapeutic areas and 5 drugs or active ingredients within each of those. The dataset is about 12,500 words strong and the POS Tags are annotated using the Penn Tree Bank part-of-speech tag-set.

**Table 2:** Un-relaxed Sentence-level POS Accuracies

| Dataset | Stanford PCFG | RAGE Hybrid POS | Diff. |
|---|---|---|---|
| **Penn Tree Bank 3** | | | |
| Token-level | 95.67 | 96.86 | 1.19 |
| Sentence-level | 31.61 | 57.91 | 26.3 |
| **RAGE Reuters 110** | | | |
| Token-level | 95.12 | 97.53 | 2.41 |
| Sentence-level | 37.71 | 63.74 | 26.03 |
| **RAGE PubMed 110** | | | |
| Token-level | 91.91 | 96.37 | 4.46 |
| Sentence-level | 25.54 | 56.62 | 31.08 |

**Table 2** provides the results of comparison of the two systems on the 3 datasets. For all the 3 datasets, token level accuracies are very high. However, as is evident, sentence level accuracies are significantly different. The Stanford PCFG has sentence-level accuracies of about ~32%. The higher numbers on the Reuters dataset than the PubMed dataset implies that the words in Reuters may have a significant semantic/structural overlap with the dataset on which PCFG was trained.

---

**Sentence 1:** Wall Street posts sharp gains, fueled by strong consumer data.

**Error:** *posts* got incorrectly identified as a Noun.

**RAGE POS Correction Rule:**
IF
  **this-phrase** is a **Noun-Phrase**
    (*Wall Street posts*) AND
  **this-phrase** contains a non-**phrase-head Named-Entity**
    (*Wall Street*) AND
  **succeeding-phrase** is a Noun-Phrase (*sharp gains*) AND
  **this-phrase-head** has a POS-wise **polysemous token** (*post*)
THEN
  Break **this-phrase** to separate the phrase-head AND
  Assign **phrase-head** a POS Tag of a **VB**.

**RAGE Rule Outcome:** Wall/NNP  Street/NNP  posts/VB  sharp/JJ  gains/NNS  ,/,  fueled/VBN  by/IN  strong/JJ  consumer/NN  data/NNS

**Sentence 2:** An accompanying record of paralanguage factors for the second example might also note a throaty rasp.

**Error:** *accompanying* for incorrectly identified as a Verb.

**RAGE POS Correction Rule:**
IF
  **this-token** is a **VBG** (*accompanying*) AND
  **prev-token** is a **DT** (*an*) AND
  **next-phrase** is a **NN\***
THEN
  Assign **this-token** a POS Tag of a **JJ**.

**RAGE Rule Outcome:** An/DT  accompanying/JJ  record/NN  of/IN  paralanguage/NN  factors/NNS  for/IN  the/DT  second/JJ  example/NN  might/MD  also/RB  note/VB  a/DT  throaty/JJ  rasp/NN  ./.

**Figure 2(a):** RAGE POS Correction Rules Outcome on *Sentence 1* and *Sentence 2* from *Fig. 1*

---

RAGE Hybrid POS Tagger consistently outperforms Stanford PCFG POS-tagger at both – sentence and token levels – in all the three datasets by ~27.8% (on an average) at sentence-levels and by ~2.7% (on an average) at token-levels. Relative absolute performance of RAGE Hybrid POS Tagger on PubMed dataset shows the robustness of the linguistic POS rules within the system. Robustness is evident from the fact that RAGE POS Rules corrected more than 30% of the errors in Stanford PCFG, just by correcting English Language inconsistencies, without any pharma domain training.

---

**Sentence 3:** New home sales jumps in the third quarter and exceeds existing home sales.

**Error:** *New Home Sales* and *Existing Home Sales* are valid concepts in the Industry.

**RAGE POS Correction Rule:**
IF
  **this-phrase** is a **Noun-Phrase**
    (*New Home Sales* and *Existing Home Sales*) AND
  **this-phrase** contains tokens which are adjectives
    (*New* and *Existing*) AND
  **this-phrase** has a sub-string with a combination
    of adjective and noun tokens is a base concept
    in the Domain Ontology
    (*New Home Sales* and *Existing Home Sales*)
THEN
  Assign adjective tokens of the phrase a **NN**.

**RAGE Rule Outcome:** New/NN  home/NN  sales/NNS  jumps/VBZ  in/IN  the/DT  third/JJ  quarter/NN  and/CC  exceeds/VBZ  existing/NN  home/NN  sales/NNS  ./.

**Sentence 4:** One thing's for sure: There have been a ton of them, and greater beings than the editors of the Wall Street.

**Error:** *Contraction* is incorrectly identified as a *Possessive ending*

**RAGE POS Correction Rule:**
IF
  **this-token** is a **POS** (*'s*) AND
  **prev-token** is a **IN** (*for*)
THEN
  Assign **this-token** a POS Tag of a **VB**.

**RAGE Rule Outcome:** One/CD  thing/NN  's/VB  for/IN  sure/JJ  :/:  There/EX  have/VBP  been/VBN  a/DT  ton/NN  of/IN  greater/JJR  beings/NNS  than/IN  the/DT  editors/NNS  of/IN  the/DT  Wall/NNS  Street/NNS  ./.

**Figure 2(b):** RAGE POS Correction Rules Outcome on *Sentence 3* and *Sentence 4* from *Fig. 1*

**Figure 2(a)** and **Figure 2(b)** shows how the linguistic rules within the RAGE Hybrid POS system correct the POS errors by any common statistical POS Tagger. The linguistic rules utilize grammatical constraints and exceptions to rectify the incorrect generalizations learnt by the statistical POS Tagging systems from the training set. Additionally, these rules (being from a closed space) are enumerated comprehensively within the RAGE Hybrid POS system, thereby providing a deterministic, complete and comprehensive framework to achieve an improved POS system.

## V. RELAXATION IN POS-TAGGING EVALUATION

In the case of shallow NLP pipelines which do not attempt deep linguistic analysis, non-critical POS errors may not be as detrimental. We also assessed relative performance of taggers by ignoring such non-critical errors. Some examples of such errors/relaxation rules for are listed in **Table 3**.

**Table 3:** POS Relaxation Rules for Evaluation

| Rule | Description |
|---|---|
| TO ⇔ IN | The infinitive 'to' can be tagged as a Preposition and vice versa. |
| WP ⇔ WDT | Wh-pronouns can be tagged as Wh-determiners and vice versa. |
| PDT ⇔ DT | Determiners can be tagged as Pre-Determiners and vice versa. |

It is important to note that some of these evaluation rules may not be linguistically correct, but do suffice in evaluation of the correctness of POS Tags to a certain extent. For example, consider the relaxation rule – *Cardinal Numbers cannot be used interchangeably with Proper Nouns*. This relaxation will not affect the correctness of the downstream NLP processing in any foreseeable manner.

We discovered over 35 relaxation rules for evaluation. Our methodology to discover these is to analyze the POS confusion matrix, followed by a case-based linguistic analysis to convert a confusion to a relaxation rule. Statistically, the application of the rules follows the Pareto principle, i.e. 20% of the rules get applied 80% of the time. This means that most of the confusion between tags are semantically more prevalent than others, and a possibility that the tagset for the language requires more differentiation.

**Table 4** shows the comparative statistics of the two POS tagging systems (baseline: Stanford PCFG) by relaxing the constraints of correctness by applying the relaxation rules.

**Table 4:** Sentence-level Relaxed POS Accuracies

| Dataset | Stanford PCFG | RAGE Hybrid POS | Diff. |
|---|---|---|---|
| **Penn Tree Bank 3** | | | |
| Token-level | 96.77 | 98.56 | 1.79 |
| Sentence-level | 56.24 | 77.90 | 21.66 |
| **RAGE Reuters 110** | | | |
| Token-level | 97.91 | 99.13 | 1.22 |
| Sentence-level | 60.52 | 83.33 | 22.81 |
| **RAGE PubMed 110** | | | |
| Token-level | 97.57 | 98.87 | 1.30 |
| Sentence-level | 64.82 | 80.72 | 15.9 |

Relaxation improves the token-level performance of both the systems by ~3.2% (Stanford with peak at 97.91%) and ~2% (RAGE with peak at 99.13%). At the sentence-level, the improvements are ~29% (Stanford, with peak at 60.52%) and ~22% (RAGE, with peak at ~83%). For pipelines that don't target deep language understanding, the RAGE Hybrid POS Tagger assigns the correct POS Tags for more than 80% of the sentences.

## VI. CONCLUSIONS

- State-of-the-art POS taggers perform poorly at a sentence-level and high token-level metrics are misleading. This realization is critical to produce robust real-world NLP applications.

- RAGE Hybrid POS Tagger, a linguistically oriented tagger, outperforms Stanford PCFG making it more usable for both shallow and deep natural language processing tasks.

- Linguistically oriented tagging systems are robust across domains and are less reliant on the underlying training set.

## ACKNOWLEDGEMENTS
We acknowledge the contributions of *Vaishali Satwase* (Sr. Linguist) and *Rajendra Rao Singh* (NLP Developer) from the Cognitive Intelligence group at RAGE Frameworks Inc., in implementing and performing this comparative study.